\documentclass{article}

    \usepackage[preprint]{neurips_2025}

\usepackage[utf8]{inputenc} 
\usepackage[T1]{fontenc}    
\usepackage{hyperref}       
\usepackage{url}            
\usepackage{booktabs}       
\usepackage{amsfonts}       
\usepackage{nicefrac}       
\usepackage{microtype}      
\usepackage{xcolor}         
\usepackage{graphicx} 
\usepackage{amsmath}
\usepackage{array}
\usepackage{multirow}
\usepackage{enumitem}
\usepackage{algorithm,algorithmicx,algpseudocode}
\usepackage{amssymb}
\newcommand{\bx}{\boldsymbol{x}}

\newcommand{\bzero}{\mathbf{0}}
\newcommand{\bI}{\mathbf{I}}
\newcommand{\grad}{\nabla}
\newcommand{\bepsilon}{{\boldsymbol{\epsilon}}}
\title{Blind-Spot Guided Diffusion for Self-supervised Real-World Denoising}

\author{%
  Shen Cheng \\
  Dexmal \\
  \And
  Haipeng Li \\
  UESTC \\
  \And
  Haibin Huang \\
  Tele AI \\
  \And
  Xiaohong Liu \\
  SJTU \\
  \And
  Shuaicheng Liu\thanks{Corresponding author} \\
  UESTC \\
}

\begin{document}

\maketitle

\begin{abstract}
In this work, we present \textbf{B}lind-\textbf{S}pot \textbf{G}uided \textbf{D}iffusion, a novel self-supervised framework for real-world image denoising. Our approach addresses two major challenges: the limitations of blind-spot networks (BSNs), which often sacrifice local detail and introduce pixel discontinuities due to spatial independence assumptions, and the difficulty of adapting diffusion models to self-supervised denoising. We propose a dual-branch diffusion framework that combines a BSN-based diffusion branch, generating semi-clean images, with a conventional diffusion branch that captures underlying noise distributions. To enable effective training without paired data, we use the BSN-based branch to guide the sampling process, capturing noise structure while preserving local details. Extensive experiments on the SIDD and DND datasets demonstrate state-of-the-art performance, establishing our method as a highly effective self-supervised solution for real-world denoising. Code and pre-trained models will be released. 
\end{abstract}
\section{Introduction}
\label{sec:intro}

As a long-standing and fundamental task, image denoising plays a crucial role in enhancing image quality for a wide range of downstream applications ~\cite{zhou2018survey,liu2019coherent}. The goal of image denoising is to recover a clean image $x$ from a noisy observation $y$, defined as
\begin{equation}
    y = x + n
\end{equation}
where $n$ represents the additive noise. This is an ill-posed problem as both the image  $x$  and noise $n$ components are unknown and difficult to disentangle. Thus, effective denoising methods require robust image priors and noise models to estimate either the image or the noise from noisy observations.


Methods for image denoising fall into multiple categories, depending on the availability of paired noisy-clean images. In the presence of paired data, denoising can be formulated as a regression task, where deep learning models are trained to minimize the difference between their predictions and the clean ground-truth images. Recent deep networks based methods have demonstrated great success for this category ~\cite{tai2017memnet,chen2017trainable,zhou2019awgn,jain2009natural,xie2012image,mao2016image,ulyanov2018deep}. However, acquiring such data at scale is resource-intensive and often impractical.

\begin{figure}[t]
    \centering
    \includegraphics[width=0.99\linewidth]{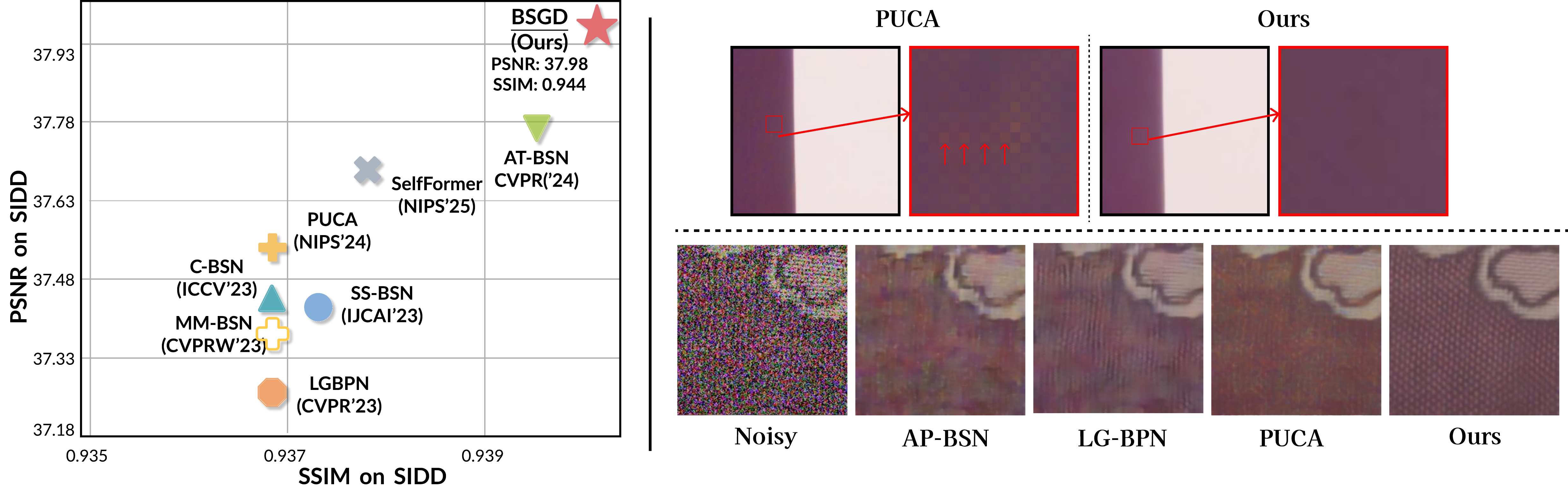}
    \vspace{-4mm}
    \caption{\textit{Left}: Comparison of different models for image denoising. Our proposed BSGD demonstrates a significant improvement over previous methods. \textit{Right Top}: Grid Pattern Effects. \textit{Right Bottom}: SIDD Validation 0035-0029. Our results presented were denoised using a guidance of $w=0.1$. All images are brightened for better visualization.}
    \label{fig:teaser}
    \vspace{-4mm}
\end{figure}

To address these challenges, self-supervised learning methods have emerged as promising alternatives by removing the dependency on paired data. Methods like Noise2Noise~\cite{lehtinen2018noise2noise} enable models to learn from unpaired noisy images by leveraging the principle that multiple noisy observations of the same underlying clean image can be used to approximate clean data. Subsequent approaches, like Noise2Void~\cite{krull2019noise2void} and Noise2Self~\cite{batson2019noise2self}, advance this paradigm by leveraging the intrinsic structure of images, allowing models to learn solely from noise using \textit{only a collection of noisy images captured in a single instance}. 

Recent advances in blind-spot networks (BSNs)~\cite{laine2019high} have extended self-supervised denoising capabilities by predicting pixel values from neighboring pixels while leaving the central pixel unseen, enabling learning from intrinsic image patterns. However, BSNs rely on an assumption of spatial noise independence, which often results in local detail loss and pixel discontinuities. Subsequent methods, such as AP-BSN~\cite{lee2022ap}, employ pixel shuffling to reduce spatial noise dependency, while networks like LG-BPN~\cite{wang2023lg}, MM-BSN~\cite{zhang2023mm}, and PUCA~\cite{jang2024puca} attempt to expand the receptive field through deeper, multi-scale features. Others, like SS-BSN and SelfFormer~\cite{quan2025selfformer}, use grid-based attention to improve contextual relationships. Nevertheless, due to pixel shuffling and blind-spot regions, these networks continue to experience losses in local details, causing pixel discontinuities, as illustrated in Fig.~\ref{fig:teaser} right.

To preserve neighboring pixel continuity, this work employs a non-blind-spot diffusion model, avoiding blind-spot masking or pixel-shuffling operations. The design is driven by two key considerations: First, diffusion models demonstrate unique advantages in distribution modeling by implicitly learning hybrid noisy distributions that jointly characterize noise patterns and latent clean image structures without explicit denoising objectives. Second, building on the theoretical framework of Classifier-Free Guidance (CFG), we propose a semi-clean guidance mechanism that dynamically adjusts score estimates during sampling, enabling directional control over generation outcomes under noisy training distributions. The synergistic integration of these properties allows diffusion models to maintain pixel coherence while achieving controllable synthesis through sampling-phase guidance.

In this paper, we propose Blind-Spot Guided Diffusion (BSGD), a novel framework that leverages the generative strength of diffusion models to overcome the limitations of BSNs. Unlike prior approaches focused on architectural modifications, we explore a new property of diffusion models for the task of denosing: the ability to combine information from both blind-spot and non-blind-spot branches during the sampling process, effectively improving denoising results. Our method employs a dual-branch diffusion framework comprising a BSN-based diffusion branch, which generates \textbf{semi-clean} images, and a conventional diffusion branch, which learns the \textbf{noisy image distribution}. Since training a conventional diffusion model without paired data is challenging, we use the BSN diffusion as guidance during sampling to capture the noise structure effectively. The two branches are jointly optimized to produce robust noise estimates for real-world denoising, leading to a significant improvement in performance, as illustrated in Fig.~\ref{fig:teaser} left. Our experiments demonstrate the effectiveness of our approach, achieving state-of-the-art results on the SIDD and DND datasets. 

\textbf{Contributions}. Our contributions can be summarized as follows:
\begin{itemize}
    \item We introduce the first dual-branch diffusion framework for self-supervised denoising, tailored for real-world applications.
    \item We combine BSN-based and conventional diffusion with blind-spot guidance and Random Replacement Sampling to overcome BSN limitations and enhance detail preservation.
    \item  Our method achieves favorable performance against state-of-the-art self-supervised denoising approaches.
\end{itemize}



\section{Related works}
\label{sec:formatting}

\noindent \textbf{Supervised Methods}
Supervised methods for image denoising rely on paired data. Some approaches focus on reducing Gaussian noise through specialized network architectures. A pioneering study~\cite{burger2012image} utilized MLPs for image denoising, achieving results comparable to BM3D~\cite{dabov2007image}. DnCNN~\cite{zhang2017beyond} demonstrated the effectiveness of CNNs in denoising Gaussian noise, inspiring numerous studies~\cite{zhang2017learning, mao2016image, ren2018dn, santhanam2017generalized, zhang2018ffdnet} that proposed various architectures.

Subsequent works~\cite{zamir2020learning, cheng2021nbnet} addressed real-world challenges by training on realistic paired noisy and clean data. However, obtaining such data is often labor-intensive. To address this, CBNet~\cite{guo2018toward} proposed estimating image noise levels and explored the generalization capabilities of Gaussian denoisers. Additionally, Zhou~\emph{et al.}~\cite{zhou2019awgn} revealed that real-world noise exhibits strong spatial correlation, and pixel-shuffle downsampling (PD) operators can disrupt this correlation, enabling networks trained solely on Gaussian noise to perform well on real-world noise.


\noindent \textbf{Self-supervised Methods}
Lehtinen~\emph{et al.}~\cite{lehtinen2018noise2noise} introduced Noise2Noise (N2N), which trains models using multiple independent noisy images of the same scenes. Subsequently, Noise2Void (N2V)~\cite{krull2019noise2void} simplified this approach by employing a blind-spot scheme, allowing denoising from single images. Since then, more variants of the blind-spot approach have emerged, including Noise2Self~\cite{batson2019noise2self}, Noise-As-Clean~\cite{xu2020noisy}, Neighbor2Neighbor~\cite{huang2021neighbor2neighbor}, Noise2Same~\cite{xie2020noise2same}, and SASL~\cite{li2023spatially}.

Instead of masking input images, some methods focus on architectures known as Blind-Spot Networks (BSN). Laine~\emph{et al.}~\cite{laine2019high} proposed masked convolution, and Wu~\emph{et al.}~\cite{wu2020unpaired} introduced stacked dilated convolution layers to integrate the blind-spot scheme within the model, improving convergence speed and performance.

However, BSN-based methods struggle with real-world noise due to its spatial correlation. To address this, AP-BSN~\cite{lee2022ap} used pixel-shuffle downsampling (PD) to disrupt this correlation and incorporated BSN for self-supervised learning. Further developments focused on designing better BSNs, such as MM-BSN~\cite{zhang2023mm}, which uses more center-masked convolutions, LGBPN~\cite{wang2023lg}, which employs dilated convolutions to expand the receptive field, SS-BSN~\cite{han2023ss}, which utilizes grid-based attention, and PUCA~\cite{jang2024puca}, which designs a patch-shuffle downsampling method. However, these structures are based on grid receptive fields, leading to grid pattern artifacts, as shown in Fig.~\ref{fig:teaser} (right-top part). In this work, we integrate BSN into diffusion models to effectively resolve these artifacts.


\noindent \textbf{Diffusion Models} 
Diffusion models~\cite{ho2020denoising} have become one of the most popular generative models due to their exceptional ability to model data distributions. They have demonstrated state-of-the-art performance across a wide range of tasks beyond image synthesis~\cite{rombach2022high,zhang2023adding}. In high-level applications, they excel in object detection~\cite{chen2023diffusiondet}, object tracking~\cite{luo2024diffusiontrack}, 3D hand mesh synthesis~\cite{xu2024handbooster}, and pose estimation~\cite{holmquist2023diffpose}. Additionally, for low-level tasks, diffusion models have shown impressive results in image alignment~\cite{li2024dmhomo, saxena2023surprising}, rectangular image stitching~\cite{zhou2024recdiffusion}, depth estimation~\cite{ke2023repurposing}, dense matching~\cite{nam2023diffusion}, and image restoration~\cite{wang2022zero, gao2023implicit, jiang2023low}.
Recently, Li~\emph{et al.}~\cite{li2024stimulating} employed Stable Diffusion~\cite{rombach2022high} for zero-shot image denoising, while Yang~\emph{et al.}~\cite{yang2024real} achieved image denoising by generating noisy images. In this work, we propose the first diffusion framework for self-supervised denoising tasks.

\section{Method}
\begin{figure*}
    \centering
    \includegraphics[width=0.99\linewidth]{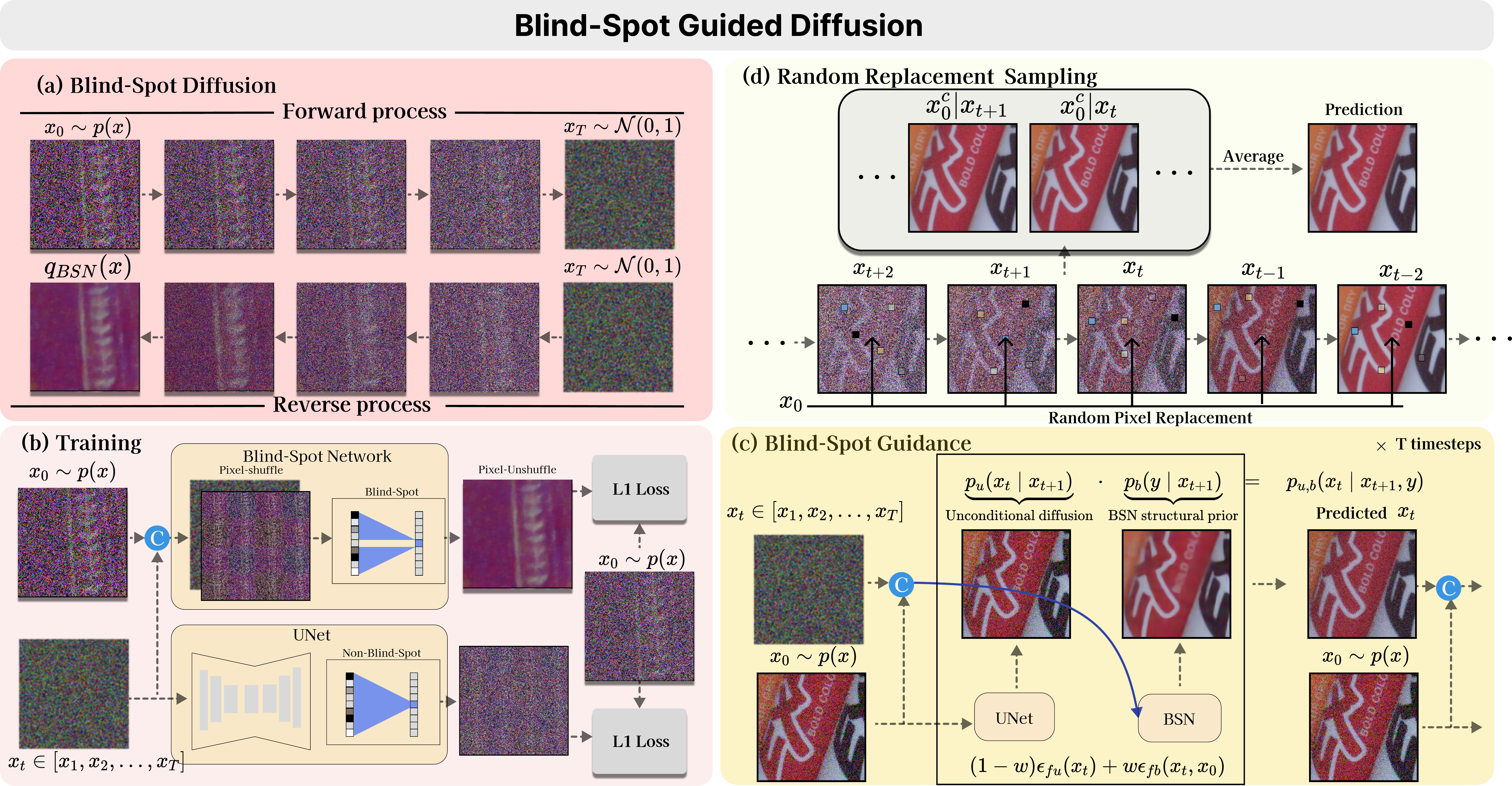}
    \caption{The Framework of Guided Blind-Spot Diffusion. (a) The training process incorporates real-world noise, while the reverse process generates clean images. (b) The framework consists of two diffusion models: one for estimating the noise-free image and the other for capturing local structures. (c) During the sampling phase, the Blind-Spot Network (BSN) serves as a guide for the UNet. (d) We implement a replacement strategy during sampling, averaging all estimated \( [\mathbf{x}_0^c | \mathbf{x}_t]_{t=0}^T \).}
    \label{fig:ppl}
    \vspace{-4mm}
\end{figure*}

The proposed framework is depicted in Fig.~\ref{fig:ppl}. Our goal is to estimate the clean distribution \( \bx^c_{0} \sim q_{0}(\bx^{c}) \) by modeling the real-world noisy data distribution \( \bx_{0} \sim p_{0}(\bx) \) and generating clean samples through the reverse diffusion process. Towards this end, we introduce the Blind-Spot Guided Diffusion (BSGD) model, which incorporates blind-spot networks. To mitigate the detail loss often seen in BSNs, we also train a standard, non-blind-spot diffusion model. During sampling, BSGD guides this non-blind-spot model to improve sample efficacy. Additionally, we implement a Complementary Replacement Sampling module to further enhance denoising performance.



\noindent \textbf{Diffusion models} 
Diffusion models (DMs) operate by learning a sequence of transformations that progressively refine a simple noise distribution into the complex distribution of real-world images—here focusing on noisy images, represented as \(\bx_{0} \sim p(\bx_{0})\).
The process initiates with a forward diffusion step, where noise is systematically introduced into the data while sampling latent variables \(\bx_{1}, \ldots, \bx_{T}\) of identical dimensionality. This step is governed by a Gaussian function that is contingent on the current state of the data and a predetermined noise schedule. The forward diffusion can be mathematically expressed as:

\begin{equation}
q(\bx_{t}|\bx_{t-1}) = \mathcal{N}(\bx_{t}; \sqrt{1 - \beta_{t}} \bx_{t-1}, \beta_{t} \mathbf{I}),
\end{equation}

\noindent where \(\beta_{t}\) adheres to a fixed schedule and \(\mathbf{I}\) denotes the identity matrix.

The sampling component of DMs involves reversing the noise accumulation process. The model learns parameters to revert the noisy data back to the image, which is also modeled using a Gaussian distribution characterized by a learned mean and a fixed variance. Following established research in diffusion models \cite{song2020score, ho2020denoising}, we adopt the generation framework:

\begin{equation}
p_{\theta}(\bx_{t-1}|\bx_{t}) = \mathcal{N}(\bx_{t-1}; \mu_{\theta}(\bx_{t}, t), \sigma_{t}^2 \mathbf{I}),
\end{equation}
where \(\mu_{\theta}\) represents the learned mean and \(\sigma_{t}^2\) is the fixed variance at time step \(t\). 

\subsection{Blind-Spot Diffusion Guidance}
Inspired by prior work~[{\color{blue}27,69}], we model the noisy observation as a perturbed version of the clean image, i.e., \( p_{0}(x) \sim N(x^{c}, \Sigma) \), where $\Sigma$ is the covariance matrix. This assumption implies that the noisy data distribution lies in a neighborhood around the clean image manifold. Diffusion models, trained on noisy images only, inherently learn to approximate the clean image distribution through iterative denoising (see supplementary). However, unconditioned diffusion sampling often produces \textit{uncontrolled} outputs, as the model lacks explicit guidance to prioritize plausible clean configurations. To address this, we draw inspiration from \textit{classifier-free guidance (CFG)}~\cite{ho2022classifier}, a standard process that trains both conditional and unconditional diffusion models and \textbf{mixes their output noise} during sampling. To achieve this, we implement two diffusion models: a conditional network utilizing a BSN architecture and an unconditional network based on U-Net. We leverage BSNs to provide a \textbf{coarse estimate} of \( x^{c} \) for CFG conditioning. Despite the limitations of BSNs, they encode structural coherence. This estimate acts as a ``soft prior" to steer the diffusion process toward plausible clean configurations. BSGD integrates this prior by reparameterizing the CFG:
\begin{equation}
\vspace{-1mm}
\footnotesize
p_{u, b}\left(x_t \mid x_{t+1}, y\right) = Z \cdot \underbrace{p_{u}\left(x_t \mid x_{t+1}\right)}_{\text{Unconditional diffusion}} \cdot \underbrace{p_{b}\left(y \mid x_t\right)}_{\text{BSN structural prior}},
\end{equation}
where \( Z \) ensures normalization. The product of these terms prioritizes regions where the diffusion model’s noise distribution aligns with the BSN’s structural prior, effectively filtering out implausible artifacts while preserving global coherence. 



\noindent\textbf{Blind-Spot Diffusion} To facilitate training within the diffusion framework, the network was adapted to be time-dependent, denoted as $f_b(\bx_t, \bx_0, t)$:
\begin{equation}
\min_{\theta} \sum_{t} \Vert f_{b}(\bx_{t}, \bx_{0}, t) - \bx_{0} \Vert_1.
\end{equation}

Due to the nature of the BSNs, once the model is trained, each estimation produced by the BSN corresponds to a clean image  \(\bx^{c}_{0|t} =  \bx^c_0 | \bx_t\). Therefore, for each sampling instance, once the first sampling \(\bx^c_0 | \bx_T\) is done, the generation framework of the blind-spot diffusion can be modified as:
\begin{equation}
p_{\theta}(\bx_{t-1}|\bx_{t}, \bx^{c}_{0|t}) = \mathcal{N}(\bx_{t-1}; \mu_{\theta}(\bx_{t}, \bx^{c}_{0|t}, t), \sigma_{t}^2 \mathbf{I}),
\end{equation}

\noindent \textbf{Non-Blind-Spot Diffusion.} The Non-blind diffusion refers to the standard diffusion framework introduced by ~\cite{ho2020denoising}, which exploits UNet architecture $f_{u}$, hence the training objective is:

\begin{equation}
\min_{\theta} \sum_{t} \Vert f_{u}(\bx_{t}, t) - \bx_{0} \Vert_1.
\end{equation}
The UNet diffusion model is utilized to effectively capture the local structure of the noise distribution.

\noindent \textbf{Blind-Spot Guidance.} The two training objectives correspond to denoising score matching ~\cite{song2020score} and can be transformed between \(\epsilon\)-prediction and \(x_0\)-prediction using the following equation:

\begin{equation}
\epsilon(\bx_t) = \frac{\bx_{t} - \sqrt{\alpha_t} \cdot f_\theta(\bx_{t}, t)}{\sqrt{1 - \alpha_t}}, \theta \in [u, b].
\end{equation}

The \(\epsilon\)-predictions for non-blind and blind-spot diffusion are denoted as \(\epsilon_{fu}\) and \(\epsilon_{fb}\), respectively. Additionally, the diffusion models can be trained jointly by randomly omitting \(\bx_0\) from the input.

Once training is complete, image samples can be generated by simulating the reverse process:

\begin{equation}
\begin{split}
    \bx_{t-1} &= \underbrace{\sqrt{1 - \alpha_{t-1} - \sigma_t^2} \cdot \epsilon(\bx_t)}_{\text{direction to } \bx_{t}} \\
    &\quad + \sqrt{\alpha_{t-1}} \underbrace{\frac{\bx_{t} - \sqrt{1 - \alpha_t} \cdot \epsilon(\bx_t)}{\sqrt{\alpha_t}}}_{\text{estimated } \bx_{0}} 
    + \sigma_{t} \mathbf{z},
    \label{eq:base_sampling}
\end{split}
\end{equation}
For each sampling step, the blind-spot network guides the non-blind U-Net for conditional generation. The original diffusion score is modified to incorporate both networks. In our work, we employed a simple linear weighting, constraining the weights within the range of $0\sim1$ to ensure stability:

\begin{equation}
\epsilon(\bx_t) = w \epsilon_{fb}(\bx_t, \bx_0) + (1-w) \epsilon_{fu}(\bx_t).
\end{equation}

Next, we describe the whole sampling process and introduce the proposed Complementary Replacement Sampling.

\subsection{Complementary Replacement Sampling}
For simplicity, we denote the latent images sampled as \([\bx_{t}]_{t=0}^T\) in Eq.~\ref{eq:base_sampling}. It is important to note that a predicted clean image corresponds to each sampling step, which we represent as \([\bx^{c}_{0} | \bx_{t}]_{t=0}^T\). These estimations serve as distinct observations of \(\bx^c_{0}\); therefore, we compute their average to obtain the final output, as illustrated in Fig.~\ref{fig:ppl}(d). 

In each sampling step, we also implement a random replacement strategy. Specifically, pixels in the predicted images \([\bx^{c}_{0} | \bx_{t}]_{t=0}^T\) are selected with a probability \(p\) and replaced with pixels from the input noisy image \(\bx_0\). The resulting images, after replacement, are then utilized as conditions for the subsequent sampling steps. We employ two types of replacements during the sampling process: \textit{Base Replacement} and \textit{Complementary Replacement}.

\begin{figure}
    \centering
    \includegraphics[width=1.0\linewidth]{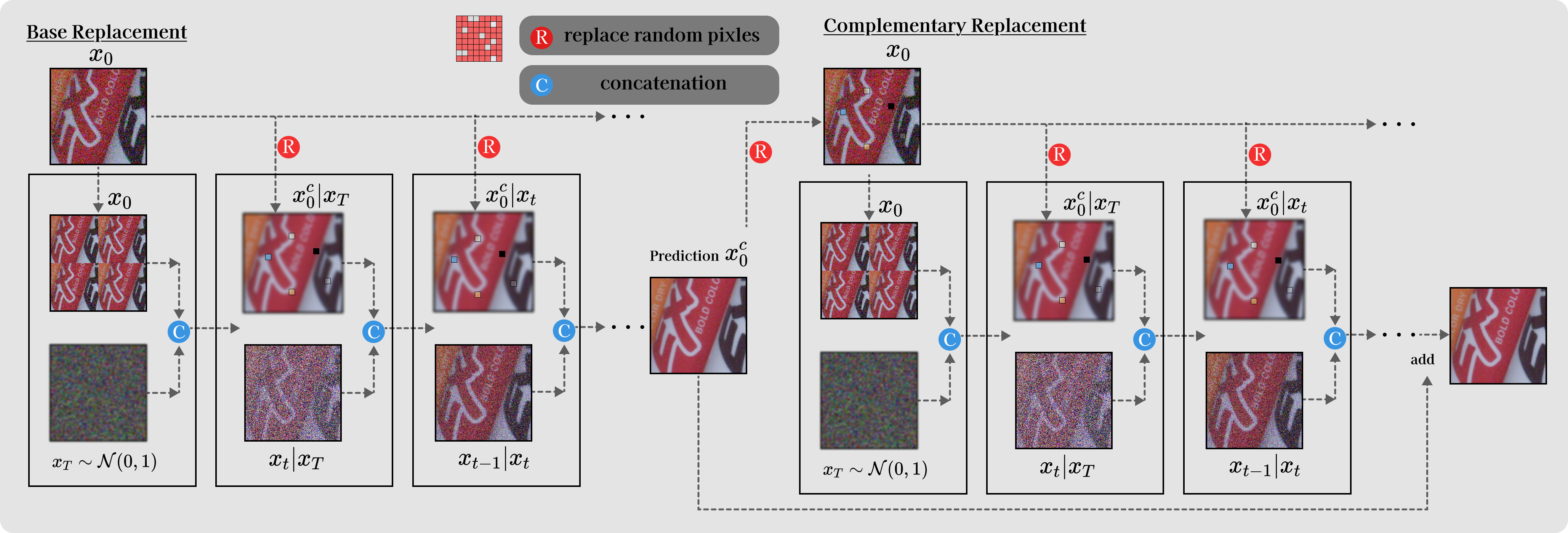}
    \caption{The Replacement Strategy. The Base Replacement is employed during the initial sampling process, while the Complementary Replacement is utilized for subsequent processes.}
    \label{fig:replacement}
    \vspace{-4mm}
\end{figure}

\noindent \textbf{Base Replacement.} The replacement process is illustrated in Fig.~\ref{fig:replacement}. The pixel-shuffling technique is employed exclusively during the first sampling step and in the context of blind-spot diffusion. In this framework, the pixel-shuffled noisy images \(\bx_0\) serve as the conditioning input, leading to the first estimation \(\bx_0^c | \bx_T\). Subsequently, this estimation undergoes the replacement process.

It is noteworthy that the sampling procedure can be conceptualized as an ordinary differential equation (ODE) ~\cite{song2020score}, allowing for sampling with larger steps. As established in ~\cite{song2020denoising}, we can derive \(\bx_t | \bx_T\) from \(\bx_0^c | \bx_T\). For the subsequent steps, the modified \(\bx_0^c | \bx_T\), post-replacement, is concatenated with the \(t\)-step sample \(\bx_t | \bx_T\) to form the input  \([\bx_0^c | \bx_T ,\bx_t | \bx_T]\)for the diffusion process. Consequently, noisy images are utilized only in the initial step, while portions of the noisy pixels are retained for the remaining steps.

\noindent \textbf{Complementary Replacement.} We define a complete sampling process as a sampling round. Upon completion of the initial sampling round, we conduct several additional sampling rounds utilizing the complementary replacement strategy. In the base replacement approach, pixels in the predicted clean image are substituted with corresponding pixels from the input noisy image. This method directly integrates noise information into the predicted output.

Conversely, the complementary replacement technique involves first substituting the input noisy image $\bx_0$ with the previously predicted clean image. This modified image then serves as the conditioning input for subsequent sampling rounds. By leveraging the previously estimated clean image, the complementary replacement aims to refine the sampling process further, potentially enhancing the quality of the final output. This dual approach, Base and Complementary Replacement, facilitates a more robust exploration of the latent space, allowing for improved convergence towards a cleaner image representation.

\begin{figure*}
    \centering
    \includegraphics[width=0.9\linewidth]{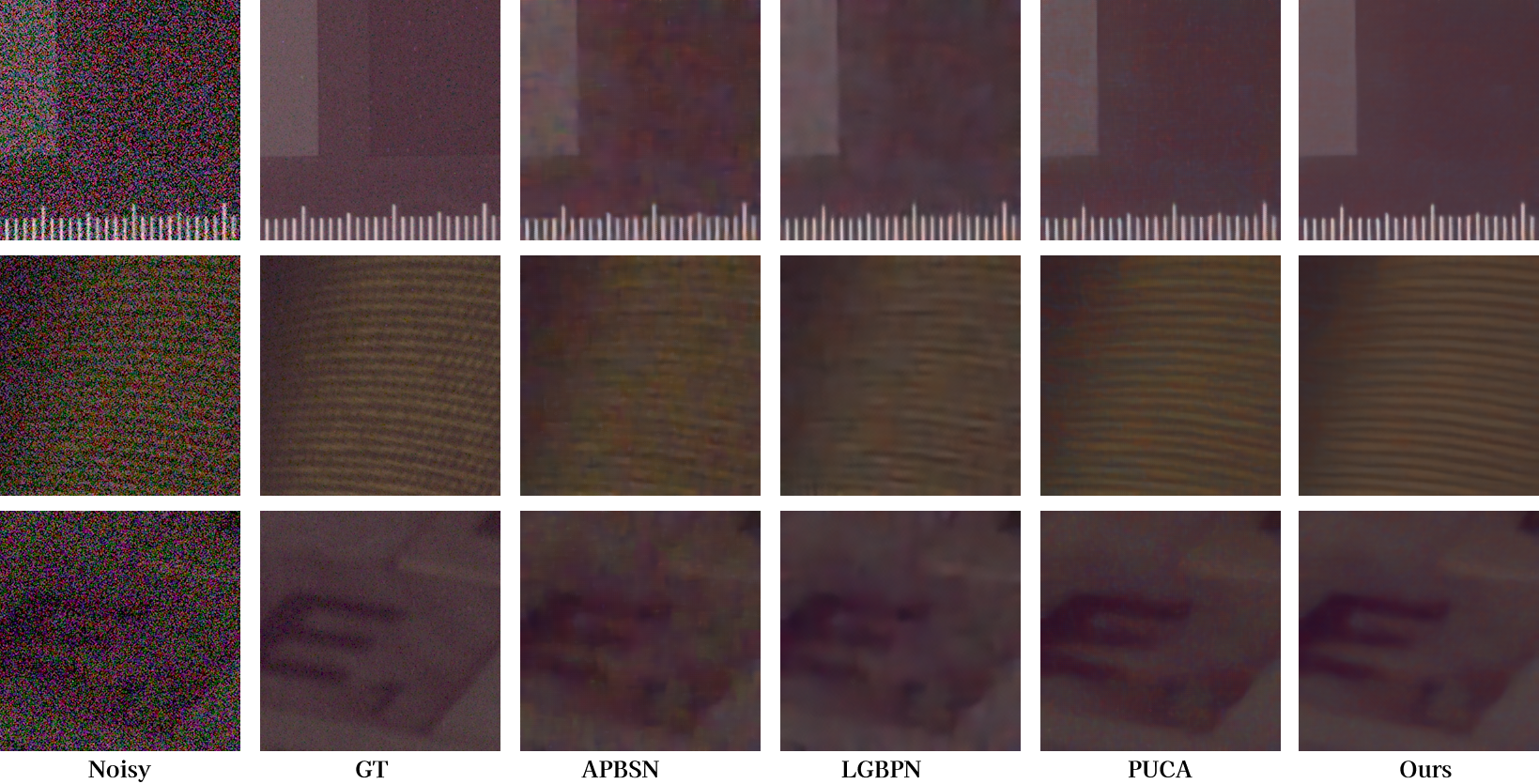}
    \vspace{-4mm}
    \caption{Denoising examples from SIDD validation dataset.~\cite{abdelhamed2018high}}
    \label{fig:vis_sidd}
    \vspace{-4mm}
\end{figure*}
\section{Experiments}

\subsection{Training Settings}

\noindent \textbf{Dataset Preparation}
For benchmarking purposes, we evaluate the proposed method through experiments conducted on the \textbf{SIDD} and \textbf{DND} datasets. The correlation between adjacent noise pixels in RAW images is weaker compared to sRGB images, as sRGB contains numerous non-linear transformations introduced by the ISP. The  Gaussian noise and \textbf{SIDD-Raw} datasets are not considered in our experiments.

The SIDD (Smartphone Image Denoising Dataset) comprises 30,000 noisy images captured across 10 distinct scenes under varying lighting conditions, utilizing five representative smartphone cameras. 
For benchmarking, SIDD extracts 1,280 color patches from 40 noisy images, which are utilized for validation and performance assessment. The benchmark results are reported on an online server, where ground-truth data is not accessible to researchers. The validation dataset, containing noisy-clean image pairs, is employed for our ablation studies. We utilize the SIDD-medium dataset for training, applying random cropping to generate image patches of size \(128 \times 128\).
\begin{table}[t]
\caption{Quantitative comparisons on benchmark datasets are presented, with bold values indicating the best performance among self-supervised methods. A dash (–) denotes instances where data is not reported. (The SIDD benchmark server (updated Oct 2024) adopted a different SSIM calculation method. However, SOTA methods in the table lack open-source code, preventing SSIM comparisons under the new framework. We thus retain original metrics while providing full experimental results to support future benchmarking of subsequent methods.)}
    \centering
    \resizebox{1.\linewidth}{!}{
    \begin{tabular}{
    >{\centering\arraybackslash}p{2.5cm}
    >{\centering\arraybackslash}p{5.0cm}||
    >{\centering\arraybackslash}p{1.8cm}
    >{\centering\arraybackslash}p{1.8cm}||
    >{\centering\arraybackslash}p{1.8cm}
    >{\centering\arraybackslash}p{1.8cm}
    }
    \toprule
        & \multirow{2}{*}{Method} & \multicolumn{2}{c||}{SIDD Benchmark} & \multicolumn{2}{c}{DND Benchmark} \\
      & & PSNR~$\textcolor{black}{\uparrow}$ & SSIM~$\textcolor{black}{\uparrow}$ & PSNR~$\textcolor{black}{\uparrow}$ & SSIM~$\textcolor{black}{\uparrow}$\\
    \hline
    \multirow{5}{*}{Supervised Real pairs} 
    & DnCNN~\cite{zhang2017beyond} & 37.64 & 0.937 & 32.43 & 0.790 \\
    & DANet~\cite{yue2020dual} & 39.47 & 0.918 & 39.59 & 0.955 \\
    & MIRNet~\cite{zamir2020learning} & 39.72 & 0.959 & 39.88 & 0.956 \\
    & NBNet~\cite{cheng2021nbnet} & 39.74 & 0.959 & 39.89 & 0.955 \\
    & NAFNet~\cite{chen2022simple} & 40.30 & 0.961 & - & - \\
    \hline
    \multirow{4}{*}{Zero-Shot} & NAC~\cite{xu2020noisy} & - & - & 36.20 & 0.925 \\
    & ScoreDVI~\cite{cheng2023score} & 34.60 & 0.920 & - & - \\
    & self2self~\cite{quan2020self2self} & 29.51 & 0.651 & - & - \\
    & MASH~\cite{chihaoui2024masked} & 34.78 & 0.900 & - & - \\
    \hline
    
    \hline
    \multirow{16}{*}{Self-supervised} & N2V~\cite{krull2019noise2void} & 27.68 & 0.668 & - & - \\
     & Noise2Self~\cite{batson2019noise2self} & 29.56 & 0.808 & - & - \\
    & \emph{Laine et al.}~\cite{huang2021neighbor2neighbor} & 30.14 & 0.823 & 35.13& 0.862- \\
    & R2R~\cite{pang2021recorrupted} & 34.78 & 0.898 & - & - \\
    & CVF-SID~\cite{neshatavar2022cvf}  & 34.71 & 0.917 & 36.50 & 0.924 \\
    & AP-BSN~\cite{lee2022ap}  & 36.91 & 0.931 & 38.09 & 0.937 \\
    & LG-BPN~\cite{wang2023lg}  & 37.28 & 0.936 & 38.43 & 0.942 \\
    & MM-BSN~\cite{zhang2023mm}  & 37.37 & 0.936 & 38.46 & 0.940 \\
    & SASL~\cite{li2023spatially}  & 37.41 & 0.934 & 38.18 & 0.938 \\
    & SS-BSN~\cite{han2023ss}  & 37.42 & 0.937 & 38.46 & 0.940 \\
    & C-BSN~\cite{fan2024complementary}  & 37.43 & 0.936 & 38.62 & 0.942 \\
    &\emph{Jang et al.}~\cite{huang2021neighbor2neighbor} & 37.53 & 0.936 & 38.56& 0.9402 \\
    & PUCA ~\cite{jang2024puca} & 37.54 & 0.936 & 38.83 & 0.942 \\
    & AT-BSN ~\cite{chen2024exploring} & 37.77 & 0.942 & 38.29 & 0.939 \\
    & SelfFormer~\cite{quan2025selfformer}  & 37.69 & 0.937 & 38.92 & 0.943 \\
    & Ours  & \textbf{37.98} & \textbf{0.944} & \textbf{38.99} &  \textbf{0.943} \\

    \hline
    \end{tabular}
    }
    \vspace{-6mm}
    
    \label{tab:benchmarks}
\end{table}

The Darmstadt Noise Dataset (DND)~\cite{plotz2017benchmarking} consists of 50 pairs of real noisy images and their corresponding ground-truth images, captured using consumer-grade cameras with varying sensor sizes. Each pair includes a source image taken at the base ISO level, while the noisy image is captured at a higher ISO with an appropriately adjusted exposure time. The reference image undergoes meticulous post-processing, which includes small camera shift adjustments, linear intensity scaling, and the removal of low-frequency bias. This post-processed image serves as the ground truth for the DND denoising benchmark. It is important to note that the DND benchmark results are also obtained by submitting predictions to the server for evaluation.

\noindent \textbf{Implementation Details} The proposed framework incorporates two distinct types of networks: a blind-spot network and a non-blind-spot UNet. The BSN architecture is derived from the PUCA model~\cite{jang2024puca}, which we have modified to create a time-dependent network utilizing a diffusion model. This modification introduces two additional inputs: the time step and the latent sampled image \(\mathbf{x}_t\). Consequently, the PUCA model serves as our baseline for comparative analysis and ablation studies.
\begin{figure*}[t]
    \centering
    \includegraphics[width=0.99\linewidth]{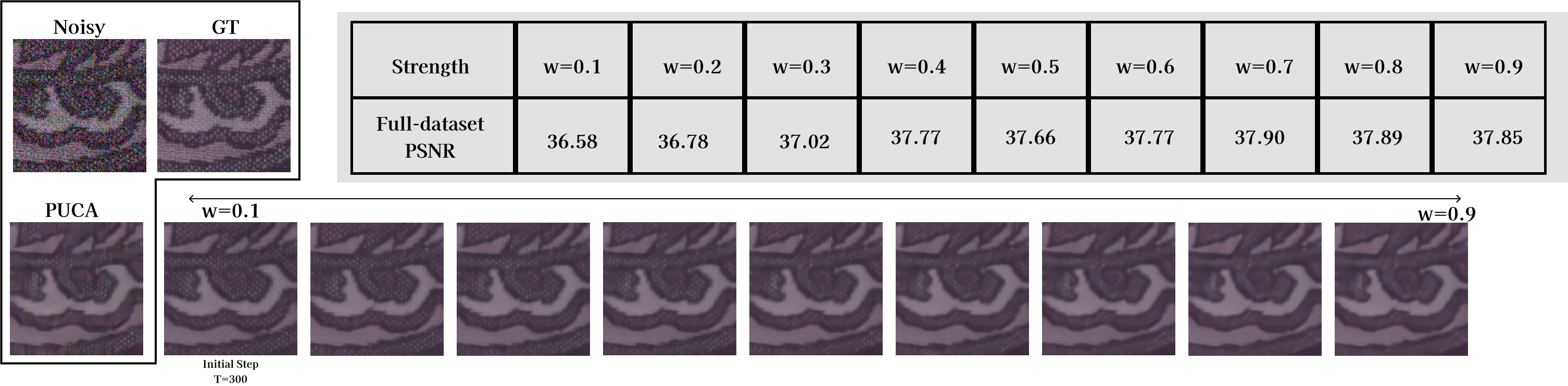}
    \vspace{-4mm}
    \caption{Ablation Study of Guidance Strength on SIDD Validation dataset. When the guidance strength \( w \) is small, finer details are preserved; however, a larger \( w \) results in blurriness. To achieve a balance, we select a higher strength for benchmarking purposes.}
    \vspace{-4mm}
    \label{fig:abl_guidance}
\end{figure*}
In contrast, the UNet architecture, widely employed in diffusion models~\cite{ho2020denoising}, features a non-blind perceptive field. We train both diffusion models under the following experimental conditions: a batch size of 8, a learning rate of \(8 \times 10^{-5}\), and the Adam optimizer. The Blind-Spot diffusion model is trained for 200,000 iterations, while the UNet diffusion model undergoes training for 500,000 iterations. The diffusion steps for training are 1,000. During the sampling phase, we implement 8 steps for each sampling process, followed by 8 additional sampling processes. For the best results of benchmarking, we start sampling with the initial step $T=300$.
\begin{table}[t]
    \caption{Ablation Study of Key Modules on SIDD Validation dataset. PUCA serves as the baseline in this analysis.}
    \vspace{-3mm}
    \small
    \centering
        

        \resizebox{1.\linewidth}{!}{
        \begin{tabular}{c|c|c|c|c|c|c}
        \hline
         Component & PUCA & PUCA ($R^3,T=8$) & PUCA ($R^3,T=64$) & Ours(w/o guidance) & Ours(w/o Replacement) & Ours \\
        \hline
        PSNR\&SSIM &  35.38/0.875 & 37.53/0.880 & 37.54/0.879 & 37.81/0.884 & 36.58/0.877 & 37.90/0.887 \\
        \hline
        \end{tabular}}
        \label{tab:abl_compo}
        \vspace{-8mm}

\end{table}
\subsection{Resutls on SIDD and DnD}
\label{sec:benchmark_result}
Table~\ref{tab:benchmarks} presents two real-world noise benchmarks, SIDD and DnD. The table compares the results of our method with those of other approaches. Self-supervised denoising methods have recently attracted significant research interest; however, their advancements over previous techniques have been relatively modest. This may be attributed to the limitations associated with the BSN and the pixel shuffle operator. In contrast, our method demonstrates a substantial enhancement compared to the SOTA methods, achieving a PSNR metric that approaches 38 dB on the SIDD dataset. Additionally, we have attained SOTA performance on the DnD benchmark.

The visual results are illustrated in Fig.~\ref{fig:vis_sidd}, where we compare the outcomes of APBSN, LGBPN, and PUCA with our results. Notably, both the APBSN and LGBPN methods exhibit significant detail loss, whereas our approach retains finer details compared to PUCA, while also achieving a lower noise level. PUCA serves as a crucial baseline for our study, as our Blind-Spot Network (BSN) is built upon its architecture with some modifications to create a time-dependent model. Our approach demonstrates an improvement of 0.43 dB over PUCA.
\subsection{Ablation Study}

\noindent \textbf{Components} As discussed in Section \ref{sec:benchmark_result}, PUCA serves as a baseline, representing a variant of our method that does not employ the diffusion model during training. Consequently, the ablation studies incorporate the replicated PUCA method trained from scratch, with performance levels comparable to those reported in \cite{jang2024puca}. The results are presented in Table~\ref{tab:abl_compo}. PUCA~\cite{jang2024puca} also employs the replacement strategy introduced in APBSN~\cite{lee2022ap}. The optimal performance of PUCA was achieved with 8 iterations of replacement, while further iterations yielded negligible improvements. The data in the fourth column demonstrate that utilizing guidance effectively integrates non-blind-spot information, leading to performance enhancement. Additionally, the data in the fifth column indicate that the proposed random replacement sampling also contributes to a notable improvement in results.
\begin{figure}
    \centering
    \includegraphics[width=1.0\linewidth]{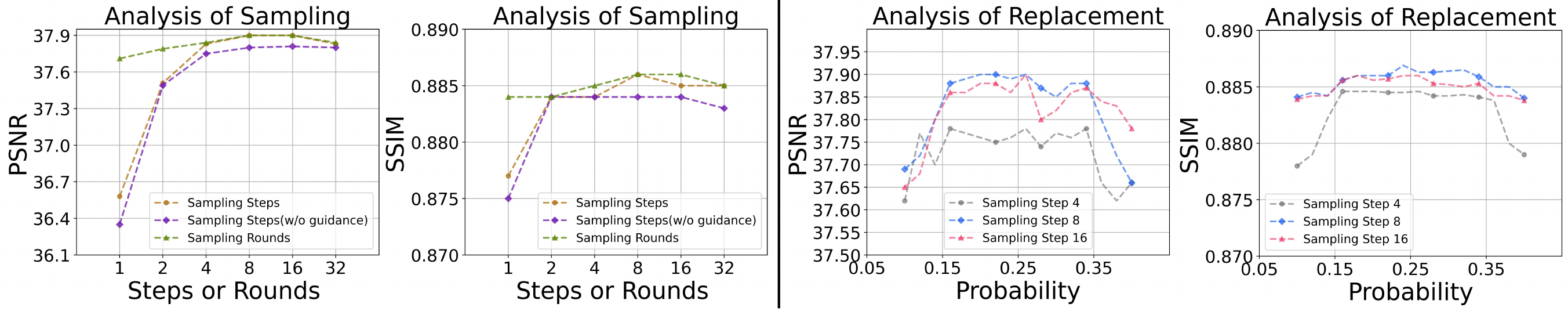}
    \vspace{-8mm}
    \caption{\textit{Left}: Analysis of Sampling Steps and Sampling Rounds. \textit{Right}: Analysis of Selection Probability}
    \label{fig:abl_sampling}
    \vspace{-8mm}
\end{figure}

\noindent \textbf{Sampling steps and Sampling Rounds}
Diffusion-based methods are fundamentally reliant on the sampling process. Given that our approach requires multiple sampling steps and complete sampling rounds, we conducted separate analytical experiments on these two critical variables.

Initially, we investigated the impact of the sampling steps, both with and without guidance. The results are summarized in Fig.~\ref{fig:abl_sampling}. As indicated in the figure, when the number of sampling steps is set to one, the performance is notably poor. This limitation arises because, at this stage, the replacement strategy cannot be effectively implemented; even with guidance, a single iteration fails to achieve satisfactory information fusion.


Performance improves significantly when steps increase to 2. Optimal results occur at 8 - 16 steps, but performance declines at 32 steps due to diffusion uncertainties. An ablation study on sampling iterations was done, with N set to 1 - 32. Results show the impact trend of iterations on performance resembles that of steps, with best results at 8 - 16 iterations and a decline at 32. The model performs well even with one sampling round as it completes the sampling process, effectively using guidance and replacement information.
The results indicate that the trend in the impact of the number of iterations on performance mirrors that of the sampling steps, with optimal results observed at 8 and 16 iterations, while performance declines at 32 iterations. Notably, unlike the sampling steps, the model demonstrates competitive performance even with just one sampling round. This is attributed to the fact that, despite having only one sampling round, the model still undergoes a complete sampling process, allowing for effective utilization of both guidance and replacement information.


\noindent \textbf{Guidance Strength} The ablation study on guidance strength is presented in Fig.~\ref{fig:abl_guidance}. We first conducted experiments on the overall performance across the entire dataset. The PSNR tends to decline when the guidance strength is close to 0 or 1, with a more pronounced decrease observed near 0. This is attributed to the predominant contribution of the unconditional non-blind-spot diffusion at that point. As the weight \( w \) increases, the PSNR gradually improves, achieving optimal results at \( w = 0.7 \) and \( w = 0.8 \), reaching a PSNR of 37.90 on the SIDD validation dataset.

Additionally, we performed visualization experiments. Notably, since sampling can commence from any step during the process, we chose to start sampling at the 300th step (out of a total of 1000 steps) to preserve image information. For example, the sampling trajectory is \([\scriptstyle 300 \to 263 \rightarrow 226 \rightarrow 189 \rightarrow 152 \rightarrow 115 \rightarrow 78 \rightarrow 41 \rightarrow 0]\) This choice ensures that even with weak guidance, random samples are not generated. It is evident that when using weak guidance, the model demonstrates strong detail recovery and denoising capabilities; however, the overall PSNR performance across the dataset is not optimal. This is due to the diffusion process introducing generative capabilities, which diminishes the reference value of the PSNR. Nevertheless, for the SIDD dataset, we selected the parameters that yielded the highest PSNR based on a compromise from our experiments.

\noindent \textbf{Replacement probability} We also conducted an experimental analysis on the probability of the replacement operation, with the results illustrated in Figure 6. Within the range of 0.15 to 0.35, the model's performance shows minimal sensitivity to the probability of selecting pixels, as both PSNR and SSIM fluctuate within a narrow band. However, performance significantly declines when the probability is below 0.15 or above 0.35. Based on this experimental analysis, we recommend selecting a probability value around 0.25 for pixel selection.



\section{Conclusion}
In this paper, we revisit the challenge of self-supervised denoising for real-world applications and introduce a first diffusion-based framework tailored for this task. Our experiments demonstrate that non-blind-spot information can be effectively integrated through diffusion sampling, overcoming limitations of blind-spot networks and enhancing denoising performance.  We hope this approach would inspire future advancements in self-supervised denoising.
\clearpage
{
    \small
    \bibliographystyle{plain}
    \bibliography{neurips_2025}
}

\clearpage
\setcounter{page}{1}
\maketitlesupplementary

\noindent \textbf{Inference time.} Our approach involves multiple sampling steps related to diffusion, as well as a multi-round sampling process. In our optimal experiments, we selected a complete sampling process consisting of 8 sampling iterations across 8 rounds, resulting in a total of 64 inference steps. Similarly, existing BSN methods, such as PUCA, also incorporate multiple inference processes. However, even when we allocate a sufficient number of inference steps to PUCA, there is no significant improvement in performance, as shown in Tab.~\ref{tab:abl_compo}. Additionally, we observe a substantial gap between current self-supervised methods and supervised approaches, indicating that the importance of enhancing performance outweighs that of model size. Conversely, even when our model performs only a single round of sampling with 8 inference steps, it significantly outperforms PUCA, as illustrated in Fig.~\ref{fig:abl_sampling}.

\noindent \textbf{Guidance.} In our approach, BSN serves as a guide to direct the non-blind-spot UNet during the sampling process. Through multiple diffusion sampling iterations, the information from both components is synergistically integrated. As shown in Fig.~\ref{fig:abl_guidance}, we observe that when the guidance is minimal, the PSNR performance across the entire dataset may be suboptimal; however, the denoised results exhibit significant detail recovery and are free of noise. This method of sampling for non-blind-spot information fusion presents a novel direction for self-supervised denoising approaches.

\noindent\textbf{More Details.} 
Our UNet-based diffusion model follows a self-supervised denoising framework without pre-trained weights. While L2 loss is standard for DMs, we opted for L1 loss for the non-BSN branch due to its very slightly better performance, which we attribute to its alignment with image denoising. In the revision, we included a comparison of L1 vs. L2 results. Additionally, we have detailed the computational costs (parameters, FLOPs) for both branches to address broader implementation concerns.

\begin{table}[h]
\vspace{-3mm}
\begin{tabular}{|>{\centering\arraybackslash}p{4cm}|>{\centering\arraybackslash}p{2.8cm}|>{\centering\arraybackslash}p{2.8cm}|>{\centering\arraybackslash}p{2.8cm}|}
\hline
     & ours-BSN & ours-UNet & SD-UNet \\
     \hline
Params(M)\&Macs(G) & 18.1/137.23    & 9.2/30.92   & 865/880 \\
\hline
\end{tabular}
\vspace{-4mm}
\end{table}

\begin{figure}[h]
    \centering
    \includegraphics[width=1.0\linewidth]{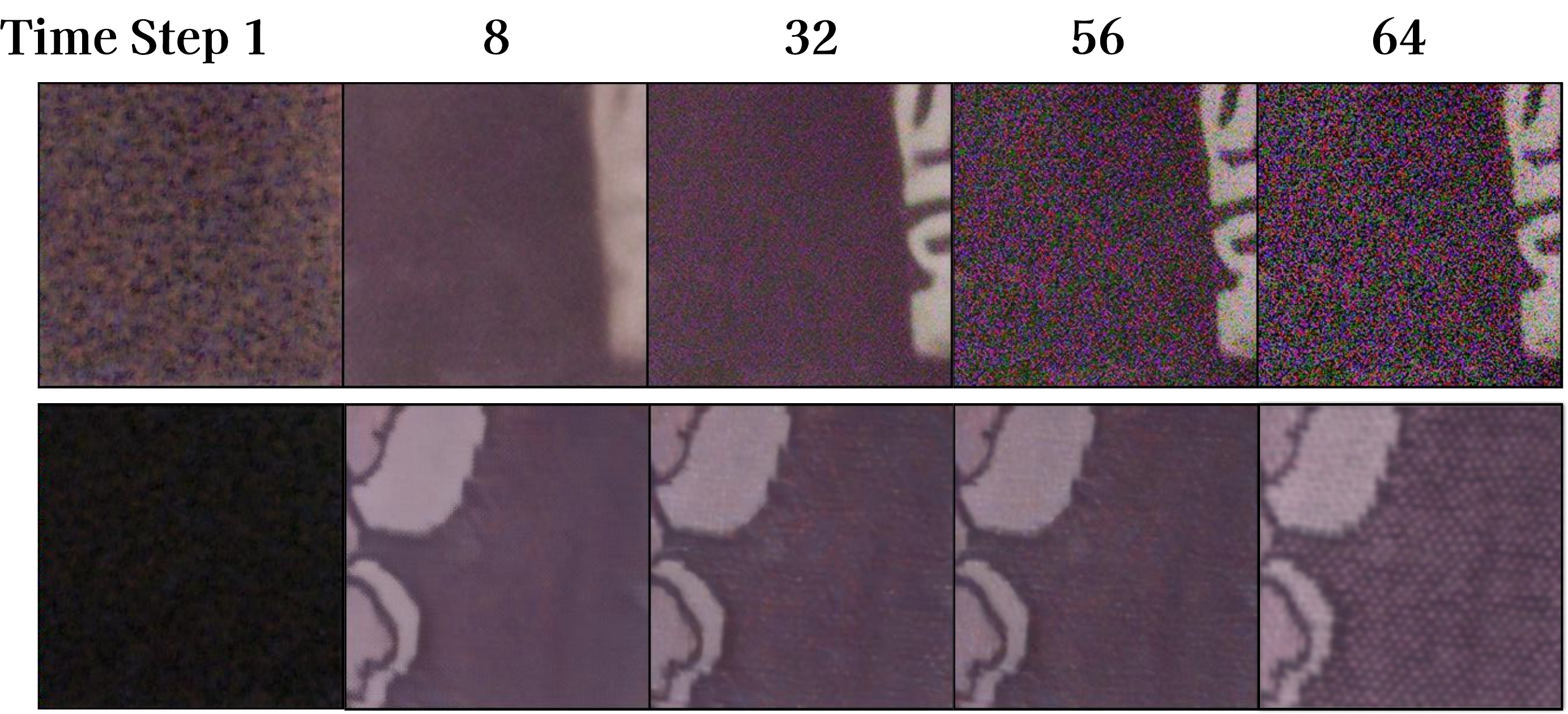}
    \vspace{-7mm}
    \caption{Output changes during the sampling process. Top: sampling without BSN prior. Bottom: sampling with BSN prior.}
    \label{fig:sampling_step}
    \vspace{-3mm}
\end{figure}

\begin{figure*}
    \centering
    \includegraphics[width=0.99\linewidth]{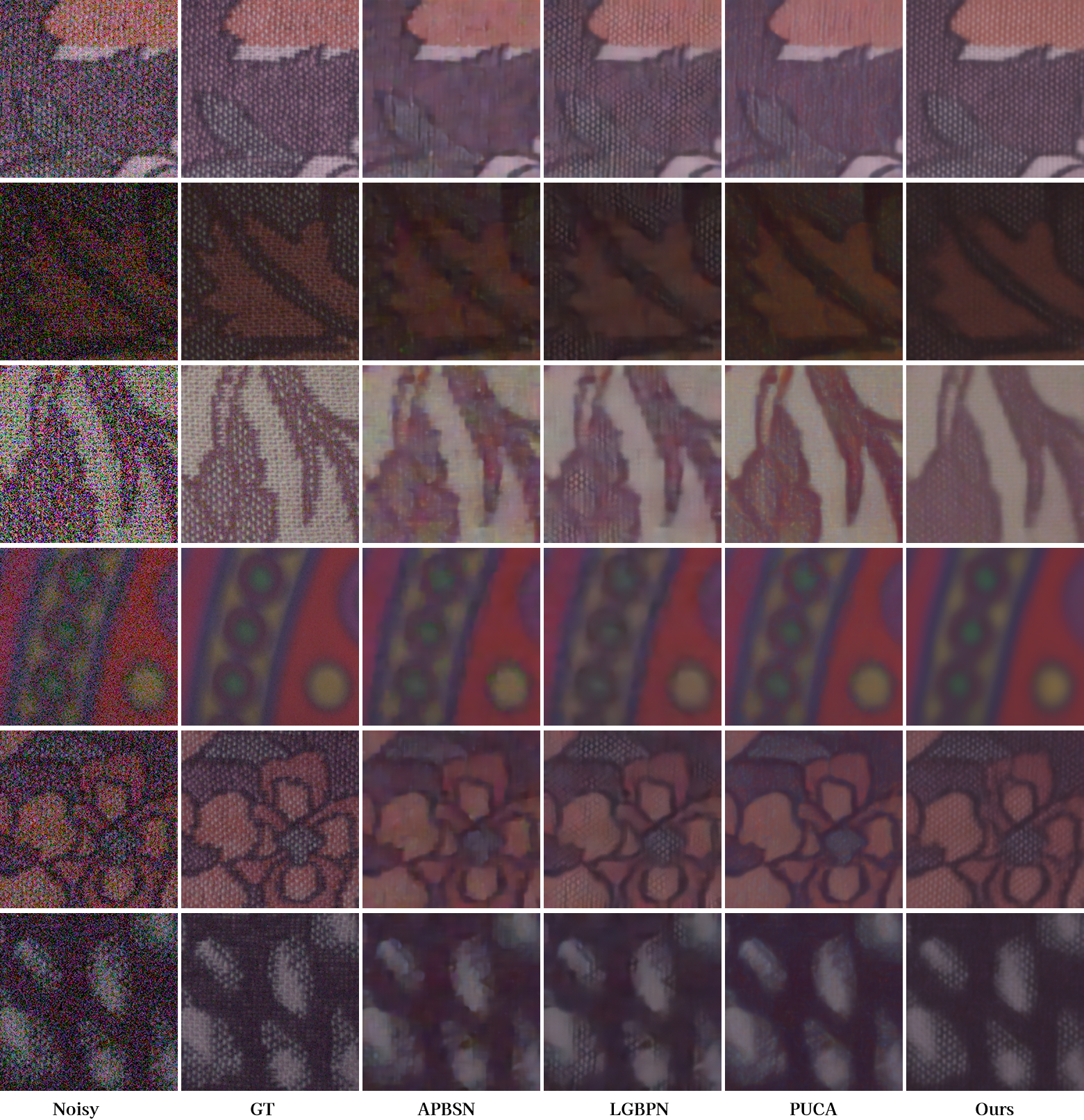}
    \caption{Our results presented were denoised using a guidance of $w=0.1$}
    \label{fig:sidd_more_results}
\end{figure*}

\noindent\textbf{More Results.} In the Fig.~\ref{fig:sidd_more_results}, we present additional subjective comparison results. These results primarily focus on images with weak texture richness. We observe that, in scenarios characterized by weak textures and relatively high signal-to-noise ratios (SNR), employing a lower guidance strength yields better outcomes. Conversely, for other types of images, a higher guidance strength is recommended.

Although we standardized the selection of a specific empirical value for the final weight $w$ in our benchmarks, tailoring the guidance strength based on the SNR could further enhance overall performance. A key insight from this work is our discovery of a method to integrate information from non-BSN structures, which may potentially eliminate the need for BSN in future applications.

\begin{algorithm}[t]
  \caption{Training} \label{alg:training}
  \small
  \begin{algorithmic}[1]
    \Repeat
      \State $\bx_0 \sim p_0(\bx)$
      \State $t \sim \mathrm{Uniform}(\{1, \dotsc, T\})$
      \State $\bepsilon\sim\mathcal{N}(\bzero,\bI)$
      \State Take gradient descent step on
      \Statex $\qquad \grad_{fu} \left\| \bx_0 - f_u(\sqrt{\bar\alpha_t} \bx_0 + \sqrt{1-\bar\alpha_t}\bepsilon, t) \right\|$
      \Statex $\qquad \grad_{fb} \left\| \bx_0 - f_b(\sqrt{\bar\alpha_t} \bx_0 +\sqrt{1-\bar\alpha_t}\bepsilon, \bx_0, t) \right\|$
    \Until{converged}
  \end{algorithmic}
\end{algorithm}
\vspace{-8mm}

\begin{algorithm}[t]
  \caption{Random Replacement Sampling} \label{alg:sampling}
  \small
  \begin{algorithmic}[1]
    \State $\bx_T \sim \mathcal{N}(\bzero, \bI)$
    \For{$t=T, \dotsc, 1$}
      \State $z \sim \mathcal{N}(\bzero, \bI)$ if $t > 1$, else $z = \bzero$
      \State pixelshuffle $\bx_t$ if $t=T$ 
      \State $\epsilon_\theta(\bx_t) = \frac{\bx_{t} - \sqrt{\alpha_t} \cdot f_\theta(\bx_{t}, t)}{\sqrt{1 - \alpha_t}}, \theta \in [u, b].$
      \State $\epsilon(\bx_t) = (1 - w) \epsilon_{fb}(\bx_t, \bx_0) + w \epsilon_{fu}(\bx_t).$
      \State $\bx_0|\bx_t= \frac{\bx_{t} - \sqrt{1 - \alpha_t} \cdot \epsilon(\bx_t)}{\sqrt{\alpha_t}}$
      \State replace $\bx_0|\bx_t$ with $\bx_0$
      \State $ \bx_{t-1} = \sqrt{1 - \alpha_{t-1} - \sigma_t^2} \cdot \epsilon(\bx_t) \quad + \sqrt{\alpha_{t-1}}\bx_0|\bx_t
    + \sigma_{t} \mathbf{z},$
        \State pixelunshuffle $\bx_t$ if $t<T$ 
    \EndFor
    \State \textbf{return} Average $\bx_0 | \bx_T, \dotsc, \bx_0|\bx_1$
  \end{algorithmic}
\end{algorithm}

\begin{algorithm}[t]
  \caption{Complementary Replacement Sampling} \label{alg:sampling}
  \small
  \begin{algorithmic}[1]
    \State $\bx_T \sim \mathcal{N}(\bzero, \bI)$
    \State $\bx_0 | \bx_R$ = Sample\_proces($\bx_T$)
    \For{$r=R-1, \dotsc, 1$}
    \State replace $\bx_0$ with $\bx_0 | \bx_R$
    \State $\bx_0 | \bx_r$ = Sample\_proces($\bx_T$)
    \EndFor
    \State \textbf{return} Average $\bx_0 | \bx_T, \dotsc, \bx_0|\bx_1$
  \end{algorithmic}
\end{algorithm}

\end{document}